\documentclass[sigconf]{acmart}

\usepackage{graphicx}
\usepackage{caption, subcaption}

\usepackage{multirow}
\usepackage[linesnumbered,ruled,vlined]{algorithm2e}

\AtBeginDocument{%
  \providecommand\BibTeX{{%
    \normalfont B\kern-0.5em{\scshape i\kern-0.25em b}\kern-0.8em\TeX}}}

\SetKwInput{KwInput}{Input}                
\SetKwInput{KwOutput}{Output}              

\setcopyright{acmlicensed}
\acmDOI{}

\acmConference[DSHealth'20]{2020 KDD Workshop on Applied Data Science for Healthcare}{August 24, 2020}{San Diego, CA, USA}
\acmBooktitle{2020 KDD Workshop on Applied Data Science for Healthcare,
 August 24, 2020, San Diego, CA, USA}
\acmPrice{}
\acmISBN{}

\begin{document}

\title{Improved Slice-wise Tumour Detection in Brain MRIs by Computing Dissimilarities between Latent Representations}


\author{Alexandra-Ioana Albu}
\affiliation{%
  \institution{Romanian Institute of Science and Technology \\  Babe\c{s}-Bolyai University}
  \streetaddress{Str.~Virgil Fulicea, 17}
  \city{Cluj-Napoca}
  \state{Romania}
  \postcode{400022}
}
\email{albu@rist.ro}

\author{Alina Enescu}
\affiliation{%
  \institution{Romanian Institute of Science and Technology \\  Babe\c{s}-Bolyai University}
  \streetaddress{Str.~Virgil Fulicea, 17}
  \city{Cluj-Napoca}
  \state{Romania}
  \postcode{400022}
}

\email{enescu@rist.ro}

\author{Luigi Malag\`o}
\affiliation{%
  \institution{Romanian Institute of Science and Technology}
  \streetaddress{Str.~Virgil Fulicea, 17}
  \city{Cluj-Napoca}
  \state{Romania}
  \postcode{400022}
}
\email{malago@rist.ro}

\renewcommand{\shortauthors}{Albu et al.}

\begin{abstract}
Anomaly detection for Magnetic Resonance Images (MRIs) can be solved with unsupervised methods by learning the distribution of healthy images and identifying anomalies as outliers. In presence of an additional dataset of unlabelled data containing also anomalies, the task can be framed as a semi-supervised task with negative and unlabelled sample points. Recently, in Albu et al., 2020, we have proposed a slice-wise semi-supervised method for tumour detection based on the computation of a dissimilarity function in the latent space of a Variational AutoEncoder, trained on unlabelled data. The dissimilarity is computed between the encoding of the image and the encoding of its reconstruction obtained through a different autoencoder trained only on healthy images. In this paper we present novel and improved results for our method, obtained by training the Variational AutoEncoders on a subset of the HCP and BRATS-2018 datasets and testing on the remaining individuals.
We show that by training the models on higher resolution images and by improving the quality of the reconstructions, we obtain results which are comparable with different baselines, which employ a single VAE trained on healthy individuals. As expected, the performance of our method increases with the size of the threshold used to determine the presence of an anomaly.
\end{abstract}





\maketitle

\section{Introduction}

Automatic outlining of anomalies in brain Magnetic Resonance Images (MRIs) is of great importance in computer-aided diagnosis, in order to develop reliable systems that can assist physicians in diagnosing pathologies.
Indeed, accurate diagnosis is not only time consuming, but also requires a lot of experience and high sensitivity to the specific condition. The study conducted by Drew et al. \cite{drew2013invisible} showed that diagnoses are vulnerable to perceptual blindness, and this can lead to high miss rates of anomalies. 

Deep learning has been employed extensively in medical imaging, with a track of very good results in the last years, see~\cite{reviewDL-MI} for a review. However, one of the limitations to the use of deep neural networks is given by the need of large datasets, which may be difficult to be obtained, especially if they need to be annotated. In the last years, this has lead to an increased interest for the design of unsupervised and semi-supervised methods. In the context of anomaly detection for medical images, a common approach consists in learning a distribution for the hidden representations of images of healthy individuals in a completely unsupervised way, and then identifying anomalies as those encodings which behave as outliers.
On the other side, there are settings as for brain tumour detection from MRIs in which datasets of unhealthy individuals are available, and they could be exploited  as unlabelled data in learning for the design of 
semi-supervised methods in presence of negative and unlabelled
samples.

In our previous work~\cite{albu|enescu|malago:2020}, we have designed a slice-wise tumour detection algorithm based on Variational AutoEncoders (VAEs) for tumour detection in brain MRIs from two publicly available datasets, HCP \cite{van2012hcp} and BRATS-2015 ~\cite{menze2014brats,kistler2013brats}. Our algorithm is based on the computation of a distance, or more generically a dissimilarity function, in the space of the approximate posteriors of a VAE, trained on both healthy (or normal) and tumoural tissues. The method requires to train an additional VAE only on healthy images, used to reconstruct unlabelled slices. The classifier is based on the computation of the distance between the encoding of an original image and the encoding of its reconstruction. 
The algorithm is semi-supervised with negative and unlabelled data.

Preliminary experiments from~\cite{albu|enescu|malago:2020} showed that the performance of our method in detecting slices which contain tumours directly depends on the quality of the reconstructions obtained with the two VAEs and in particular on the capability of the first VAE, trained only on healthy images, to remove the tumoural tissues in the reconstructed image. In this paper we present novel and improved results in which we evaluate our algorithm on the HCP and BRATS-2018 datasets, using higher resolution images and models able to produce better reconstructions.

VAEs are powerful generative models with probabilistic latent variables which have been successfully employed in various segmentation or anomaly detection tasks in medical applications \cite{tashiro2017deep, myronenko20183d, zimmerer2018context, baur2018deep}. Generative models are particularly well suited for anomaly detection, as they can be used to learn the distribution of healthy tissues and then identify anomalies as out-of-distribution behaviours.

Unsupervised or semi-supervised anomaly detection methods typically use various types of autoencoders trained on healthy data, relying on the fact that anomalies, which were not seen during training, cannot be properly reconstructed.
The encoder-decoder architecture is particularly useful as the anomaly segmentation can be easily obtained as a pixel-wise difference between the original image and its reconstruction  \cite{baur2018deep, chen2018unsupervised, chen2018deep}. Chen et~al.~\cite{chen2018unsupervised, chen2018deep} evaluated two generative models, VAEs and Adversarial AutoEncoders, for the task of detection of tumours and stroke lesions in brain MRIs. Additionally, ~\cite{chen2018unsupervised} proposes the use of a regularizer that encourages the models to learn more suitable latent space representations for healthy and unhealthy images.
In~\cite{zimmerercase} Zimmerer et~al. highlighted the limitations of using a plain reconstruction error for detecting anomalous pixels and proposed instead to quantify the abnormality of input pixels using the derivative of the log-likelihood with respect to the inputs.
In~\cite{zimmerer2018context} a context-encoding regularizer was added to the VAEs loss function, obtaining a Context Encoding VAE (ceVAE) which has shown significant improvements compared to previously proposed methods. Another approach combining the benefits of an autoencoder architecture and adversarial training, the Adversarial Dual Autoencoder (ADAE), was proposed in \cite{vu2019anomaly}. 

The paper is organized as follows. In Section~\ref{methods} we review our algorithm for anomaly detection based on the computation of dissimilarities between latent representations of a Variational AutoEncoder. In Section~\ref{experiments} we present our experimental setting, while Section~\ref{Results} shows that our method can discriminate between images containing tumours and healthy scans, and compares favourably with different baselines which employ a single VAE trained on healthy individuals. Finally, Section~\ref{conclusions} outlines the conclusions and future works.

\section{Anomaly Detection Algorithm}
\label{methods}

In this section we briefly present our algorithm for slice-wise anomaly detection in brain MRIs based on Variational AutoEncoders (VAEs), firstly introduced in ~\cite{albu|enescu|malago:2020}. The algorithm is based on the use of VAEs~\cite{kingma2013auto,rezende2014}, i.e., generative models which consist of two neural networks: an encoder and a decoder. The encoder maps the input to the parameters $\theta$ of a probability density function $q_\theta$ over the latent space, while the decoder maps the latent representation to a probability density function $p_\phi$ over the space of the observations. VAEs are trained by maximizing a lower bound for the log-likelihood, which consists of a reconstruction error and a Kullback-Leibler penalty term. For each input $x$, a VAE computes an approximate posterior from which the latent variables are sampled.

Our algorithm requires to train two VAEs on two different datasets, one containing only healthy subjects, \texttt{VAE-H}, and one containing brain scans that may or may not contain tumoural tissue,  \texttt{VAE}. The proposed method is presented in Algorithm \ref{alg:classif}. Differently from our previous approach, in order to obtain sharper reconstructions for high resolution images, we trained the VAE models by maximizing the likelihood only for pixels inside the brain mask. By using this technique, the model is able to focus on reconstructing finer brain structures details, however, background pixels can take arbitrary values. In order to obtain the original black background, the brain mask was overlapped on the reconstructed image.

To test the anomaly detection, we reconstruct a given image through the first model, \texttt{VAE-H} and compute the distance between the encondings through the second model \texttt{VAE} of the given image and its reconstruction. Due to the fact that the first model is trained only on normal tissues, it will not be able to reconstruct accurately anomalies, thus the distance between the two encodings will be larger for images containing abnormal regions.
 
\begin{small}
\begin{algorithm}[!h]
\DontPrintSemicolon
  \KwInput{Let $d$ be a distance defined over the latent space of a VAE}
  \KwData{\texttt{MRI-H}: dataset of MRI slices of healthy individuals}
  \KwData{\texttt{MRI}: dataset of MRI slices of individuals which may have tumoural tissues}
  %
Let \texttt{VAE-H} be a VAE trained on \texttt{MRI-H}\\
Let \texttt{VAE} be a VAE trained on \texttt{MRI}\\
Let $x_h$ = \texttt{VAE-H.rec}($x$) be the reconstruction of $x$ through \texttt{VAE-H}\\
Let \texttt{VAE.enc}($x$) be the encoding of $x$ through \texttt{VAE}\\
Let $d_x = d($\texttt{VAE.enc}($x$),\texttt{VAE.enc}($x_h$)$)$\\
Compute the distribution of the distances $d_x$, with $x$ in the validation set of $\texttt{MRI-H}$ and let $d_*$ be a threshold selected based on a percentile \\
Compute $d_x$, with $x$ in the test set of \texttt{MRI}, and classify the slice as healthy if $d_{x} < d_*$
\caption{Classification of brain MRIs from~\cite{albu|enescu|malago:2020}}
\label{alg:classif}
\end{algorithm}
\end{small}

\section{Experimental Setting}\label{experiments}

This section presents an overview the two datasets used in our experiments, alongside the methods used for preprocessing the data. Afterwards, the architectures of the networks and aspects related to the training procedure are outlined.

\subsection{Datasets}
Compared to our previous experiments~\cite{albu|enescu|malago:2020}, we used the same dataset of healthy individuals, i.e., the HCP dataset~\cite{van2012hcp}, but we evaluated our method on a newer version of the BRATS dataset, BRATS-2018~\cite{menze2014brats,kistler2013brats}. 
We used a subset of the HCP dataset which contains 214 T2-weighted MRI scans of unrelated subjects. We removed the black slices keeping only 190 slices containing brain tissue from the HCP dataset, and we kept the middle 130 slices from the BRATS dataset. Both datasets were split into training, validation and test datasets, each representing $70 \%, 15\%$ and $15\%$ of the total amount of data.

The images have been preprocessed by applying bias field correction using the N4ITK algorithm~\cite{journals/tmi/TustisonACZEYG10}, a variant of the popular nonparametric nonuniform intensity normalization (N3) algorithm, implemented in ANTSs~\cite{avants2009advanced}.
Next, the 3D scans of both datasets have been normalized each in the [-1,+1] range, and subsequently the histogram of each individual has been matched to a reference histogram associated to an image from the BRATS dataset. We have chosen to normalize with respect to an image coming from the BRATS dataset since images have a larger variability compared to the HCP dataset due to the presence of tumoural tissues. 

The images from both datasets have been cropped and resized to \( 200 \times 200 \) and then down-sampled to \( 128 \times 128 \) pixels. Furthermore, to avoid overfitting, the left and right hemispheres have been flipped with probability 0.5 during training and adjustment of brightness was applied, with delta 0.1 and probability 0.3. We further applied Gaussian white noise to the images given in input to the network. 

\subsection{Network Architectures}
We have trained two VAEs with the same network topology. The encoders are represented by a sequence of convolutions with channels $[64, 128, 256, 512]$, kernel size $4 \times 4$ and strides $2$, followed by a stochastic layer of independent Gaussian distributions of dimension $128$. The decoders are built symmetrically, with output channels $[512, 256, 128, 64]$, kernel size $4 \times 4$ and strides $2$. Similarly to our previous work, the output model is represented by a logit-normal distribution, for which we used a clip value for the mean equal to $0.01$ for HCP and $0.001$ for BRATS, a scalar covariance for HCP and vectorial for BRATS, with a minimum value of $0.01$. 
When training the VAEs, the expectation is computed by generating $3$ samples in the latent space for each input. The models have been trained using the Adam optimizer for 100 epochs, with a learning rate of $0.0001$ and  batch size of $32$.

\section{Results}
\label{Results}

In this section we present novel and improved results, obtained by evaluating our anomaly detection algorithm on higher resolution images, using more powerful models able to provide better reconstructions.

As in \cite{albu|enescu|malago:2020}, we have chosen in Algorithm 1 the distance function $d$ to be the norm of the difference between the means obtained with $\texttt{VAE}$ for the encodings of $x$ and $x_h$. 
The threshold $d_*$ used to classify images from the test set of BRATS has been computed as a percentile of the distances computed on the validation set of HCP. Differently from our previous work, where we fixed the $99\%$ percentile, here we selected the percentile which maximizes the F1 score on the validation set of BRATS. In addition, we performed a $5$-fold cross-validation scheme on the validation and test sets, by using $1/5$ of the data to select the percentile and computing the performance metrics on the other $4/5$ of the data.

We have used as anomaly scores for the computation of the ROC-AUC the distances in the latent space of \texttt{VAE} between the encoding of the original image and the encoding of its reconstruction through \texttt{VAE-H}. To validate our results, we have compared our method with different baselines using only the VAE trained on healthy images. These baselines are given by computing the $L^2$ norm of the difference between the original image and its reconstruction using $\texttt{VAE-H}$ for different models.

Due to the resizing of the images, the area of a lesion is being reduced and thus small tumours may be difficult to be detected. To assess the impact of the tumour size on the performance, we evaluated our method for various numbers of anomalous pixels above which we consider slices to be unhealthy.  

Our results, averaged over 5 splits and the corresponding standard deviations are presented in Table \ref{table_results} for both our method and several $L^2$ baselines trained only on healthy images. We considered a plain VAE, a denoising VAE with randomly placed grayscale square masks of size 40 in input (CE DVAE), and a VAE regularized by the $L^2$ distance between clean images and their
reconstructions in presence of randomly placed grayscale square masks of size 20 (VAE + CE reg.), cf.~\cite{zimmerer2018context}. 
The size of the masks are those which provided the best performance. Different thresholds for the tumour size have been considered. We can see that our method outperforms the baselines considering all metrics and tumour sizes. As expected, the performance increases with the threshold used to determine the presence of an anomaly.

\begin{figure}[htbp]
\begin{subfigure}[t]{1.0\linewidth}
    \centering
    \includegraphics[width=.8\textwidth]{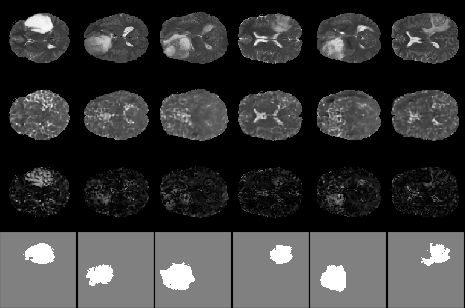}
\end{subfigure} 
\vfil
\begin{subfigure}[t]{1.0\linewidth}
    \centering
    \includegraphics[width=.8\textwidth]{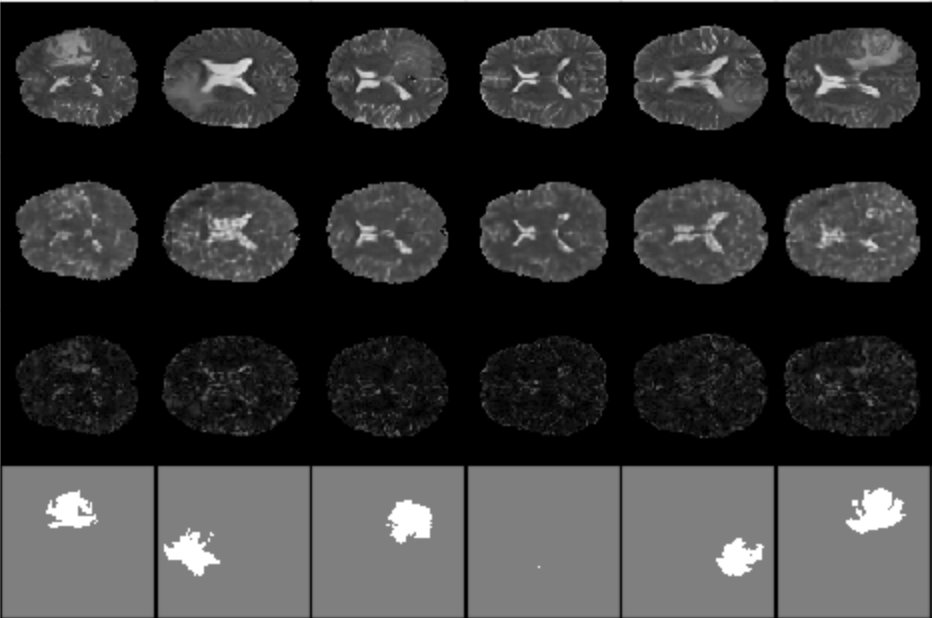}
\end{subfigure}    
    \caption{First row: original BRATS images from the test set. Second row: reconstruction through \texttt{VAE-H} of the original images. Third row: residual between original and reconstructed images. Fourth row: mask of the tumour. The two panels include cherry-picked slices to see the quality of the reconstructions and the associated residual images in different conditions. \label{fig:reconstructions-BRATS}}
\end{figure}

We further depict in Table \ref{tab:res} a comparison of our approach with state-of-the-art unsupervised anomaly detection methods. Notice that a direct comparison is not possible, because of the different number of slices in the test set and the different version of the BRATS dataset. Compared to the two related approaches, we used a higher resolution for the images and a smaller test set, since we used part of BRATS dataset in training. We report the results for a tumour size threshold of $20$, as in \cite{zimmerer2018context}.
To evaluate the capability of our \texttt{VAE-H} model to remove the tumours from the BRATS images, we report some examples in Figure~\ref{fig:reconstructions-BRATS}.

\begin{footnotesize}
\begin{table}[h]
\caption{Area Under the ROC Curve, accuracy and F1 score for BRATS-2018 (test set, averaged over 5 folds) for $L^2$ distances computed in the input space for different VAEs versus $L^2$ of the means in the latent space (our method~\cite{albu|enescu|malago:2020}). \textit{Th.} denotes the threshold expressed in annotated pixels (before rescaling) used to determine if a slice contains anomalies.
}
\label{table_results}
\begin{center}
\begin{tabular}{|c|c|c|c|c|}
\hline
Th. & Method & ROC-AUC & Accuracy & F1 score \\ \hline
\multirow{4}{*}{0} & $L^2$ input VAE & $0.848 \pm 0.004$ & $0.652 \pm 0.008$ & $0.507 \pm 0.006$ \\ \cline{2-5} 
 & $L^2$ input CE DVAE  & $0.864 \pm 0.004$ & $0.656 \pm 0.008$ & $0.505 \pm 0.007$ \\ \cline{2-5} 
 & $L^2$ input VAE + CE reg.  & $0.865 \pm 0.003$ & $0.736 \pm 0.003$ & $0.679 \pm 0.004$ \\ \cline{2-5} 
 & Our method & $\textbf{0.884} \pm 0.003$ & $\textbf{0.805} \pm 0.002$ & $\textbf{0.803} \pm 0.004$ \\ \hline
 \hline
 \multirow{4}{*}{20} & $L^2$ input VAE & $0.851 \pm 0.004$ & $0.673 \pm 0.007$ & $0.520 \pm 0.005$\\ \cline{2-5} 
 &  $L^2$ input CE DVAE  & $0.868 \pm 0.004$ & $0.678 \pm 0.007$ & $0.521 \pm 0.007$ \\ \cline{2-5} 
  & $L^2$ input VAE + CE reg.  & $0.869 \pm 0.003$ & $0.753 \pm 0.003$ & $0.691 \pm 0.004$ \\ \cline{2-5}
 & Our method & $\textbf{0.890} \pm 0.004$ &  $\textbf{0.809} \pm 0.002$ & $\textbf{0.803} \pm 0.005$ \\ \hline
 \hline
\multirow{4}{*}{50} & $L^2$ input VAE & $0.852 \pm 0.004$ & $0.684 \pm 0.006$ & $0.526 \pm 0.005$ \\ \cline{2-5} 
 &  $L^2$ input CE DVAE  & $0.871 \pm 0.004$ & $0.691 \pm 0.006$ & $0.531 \pm 0.007$ \\ \cline{2-5} 
  & $L^2$ input VAE + CE reg.  & $0.873 \pm 0.003$ & $0.762 \pm 0.002$ & $0.698 \pm 0.004$ \\ \cline{2-5}
 & Our method &  $\textbf{0.892} \pm 0.003$ & $\textbf{0.811} \pm 0.002$ & $\textbf{0.802} \pm 0.004$ \\ \hline
 \hline
 \multirow{4}{*}{150} & $L^2$ input VAE & $0.856 \pm 0.004$ & $0.707 \pm 0.005$ & $0.541 \pm 0.005$ \\ \cline{2-5} 
 &  $L^2$ input CE DVAE & $0.876 \pm 0.004$ & $0.717 \pm 0.006$ & $0.550 \pm 0.007$ \\ \cline{2-5} 
  & $L^2$ input VAE + CE reg. & $0.879 \pm 0.002$ & $0.781 \pm 0.002$ & $0.711 \pm 0.004$ \\ \cline{2-5}
 & Our method & $\textbf{0.897} \pm 0.004$ & $\textbf{0.814} \pm 0.002$ & $\textbf{0.795} \pm 0.005$ \\
 \hline
\end{tabular}
\end{center}

\end{table}
\end{footnotesize}

\begin{footnotesize}
\begin{table}[!h]
    \centering
    \caption{ROC-AUC for state-of-the-art and for our method. \textit{Res.} denotes the 
    resolution of images after rescaling, and \textit{Th.} the threshold expressed in annotated pixels (before rescaling) used to determine if a slice contains anomalies.}
    \label{tab:res}
    \begin{tabular}{|c|c|c|c|c|c|}
        \hline
        Method &  ROC-AUC & Dataset & \begin{tabular}[c]{@{}c@{}} \# patients \\ test set \end{tabular}   & Res. & Th.\\
        \hline
        ADAE \cite{vu2019anomaly} & 0.892 & BRATS2017 & 285 & 32x32 & ? \\
         \hline
        ceVAE \cite{zimmerer2018context} & 0.867 & BRATS2017 & 266 & 64x64 & 20\\
        \hline
         \hline
         Our method \cite{albu|enescu|malago:2020} & 0.890 $\pm$ 0.004  & BRATS2018 & 69 & 128x128 & 20\\
         \hline
    \end{tabular}
\end{table}
\end{footnotesize}

\section{Conclusions}
\label{conclusions}

In this paper, we have presented novel and improved results on anomaly detection for brain MRIs based on two Variational AutoEncoders (VAEs). Compared to our previous results \cite{albu|enescu|malago:2020}, several changes have been made that have conducted to an improvement of the performance. A newer version of the BRATS dataset (BRATS-2018) has been used in training, together with a larger subset of the HCP dataset containing 214 individuals. Two additional steps in pre-processing have been introduced for both datasets: bias field correction using the N4ITK algorithm and matching of the histogram for each individual, using a reference histogram associated to an image from the BRATS dataset. In order to obtain sharper reconstructions, we trained the VAE models by maximizing the likelihood only for the pixels in the brain segmentation masks, which we have pre-computed. Additionally the choice of the best threshold has been obtained using a cross validation procedure, by removing the need to fix this parameter a priori.

We are currently investigating the possibility of improving the anomaly detection performance by using a perceptual loss which has been proved to increase the quality of the reconstructions~\cite{percep_loss_tutorial} in other medical imaging datasets.

\begin{acks}
The authors are supported by the DeepRiemann project, co-funded by the European Regional Development Fund and the Romanian Government through the Competitiveness Operational Programme 2014-2020, project ID P\_37\_714,  contract no.~136/27.09.2016, SMIS code 103321. Data used in the preparation of this work were obtained from the Human Connectome Project (HCP) database\footnote{https://ida.loni.usc.edu/login.jsp} and the Multimodal Brain Tumor Segmentation Challenge
(BRATS)\footnote{https://www.med.upenn.edu/sbia/brats2018/data.html}.
\end{acks}

\bibliographystyle{ACM-Reference-Format}
\bibliography{acmart.bib}

\end{document}